\documentclass[runningheads, orivec]{llncs}

\usepackage[T1]{fontenc}
\usepackage{graphicx}
\usepackage{amsmath}
\usepackage{multirow}
\usepackage{bbding}
\usepackage{iftex}
\ifPDFTeX
  \usepackage{CJKutf8}
  \newcommand*{\Ja}[1]{\begin{CJK}{UTF8}{ipxm}#1\end{CJK}}
\else
  \usepackage{fontspec}
  \newfontfamily\japanesefont{HaranoAjiMincho-Regular.otf}
  \newcommand*{\Ja}[1]{{\japanesefont #1}}
\fi
\begin{document}

\title{Japanese Handwriting Generation \\ Based on Human Preference Feedback \\ Using Direct Preference Optimization}

\titlerunning{Japanese Handwriting Generation Based on Human Preference Feedback}

\author{
Wakana~Kuwata\inst{1}\Envelope\orcidID{0009-0009-4612-4467} \and \\
Ryota~Mibayashi\inst{2}\orcidID{0009-0005-8995-5579} \and \\
Masanori~Tani\inst{3}\orcidID{0009-0008-9433-893X} \and \\
Hiroaki~Ohshima\inst{1}\orcidID{0000-0002-9492-2246}
}

\authorrunning{W. Kuwata et al.}
\institute{
University of Hyogo, Hyogo, Japan
\email{af25x004@guh.u-hyogo.ac.jp, ohshima@ai.u-hyogo.ac.jp}
\and
Kobe University, Hyogo, Japan
\email{mibayashi@people.kobe-u.ac.jp}
\and
ACADEMIC VISION LLC, Kyoto, Japan
}

\maketitle

\begin{abstract}
We propose a Japanese handwriting generation framework that integrates human preference feedback using Direct Preference Optimization (DPO) to align generative models with human cognitive and stylistic preferences. While font generation technologies leveraging Generative Adversarial Networks and diffusion models have advanced significantly, applying these approaches to handwriting generation presents unique challenges.
Font generation is a deterministic process yielding consistent styles. Handwriting generation, on the other hand, is inherently non-deterministic. In handwriting, glyph shapes never match exactly even when the same writer writes the same character. We define this shape difference as handwriting variability. Although this variability is essential for conveying the natural quality of human writing, it complicates model training and substantially increases data collection costs to capture diverse distribution patterns.
To address these limitations, we incorporate human perceptual feedback into the training loop of handwriting generation. While collecting extensive handwritten data is costly, evaluating pairwise preferences between two generated samples remains very easy for humans. Applying DPO, an efficient RL-free preference optimization method, to our generation model enables high-quality handwriting generation without relying on large-scale dataset collection.
Experimental results across four distinct Japanese handwriting styles demonstrate that DPO-based training enhances overall structural accuracy. However, in highly complex styles with significant variability, such as brush pen writing, the model fails to capture style-specific features. These findings highlight a critical challenge in extending DPO to complex handwriting styles.
\end{abstract}

\section{Introduction}\label{sec:intro}

Glyph generation research has historically focused on font generation. Various frameworks leveraging generative models, such as Generative Adversarial Networks (GANs) and diffusion models, have been proposed to synthesize high-quality fonts~\cite{2021:ICMR:MLFont,2022:CVPR:FsFont,2017:zi2zi,2024:AAAI:FontDiffuser}.

Handwriting generation differs fundamentally from font generation. To clarify this distinction, we define the key components of characters. Linguistic information representing the character class (e.g., ``A'' or ``\Ja{あ}'') is referred to as \textbf{content}. The unique design attributes that materialize the content are defined as \textbf{style}. A \textbf{glyph} represents the physical rendering of a character, combining both content and style.

\begin{figure}[tb]
\begin{center}
\includegraphics[width=\linewidth]{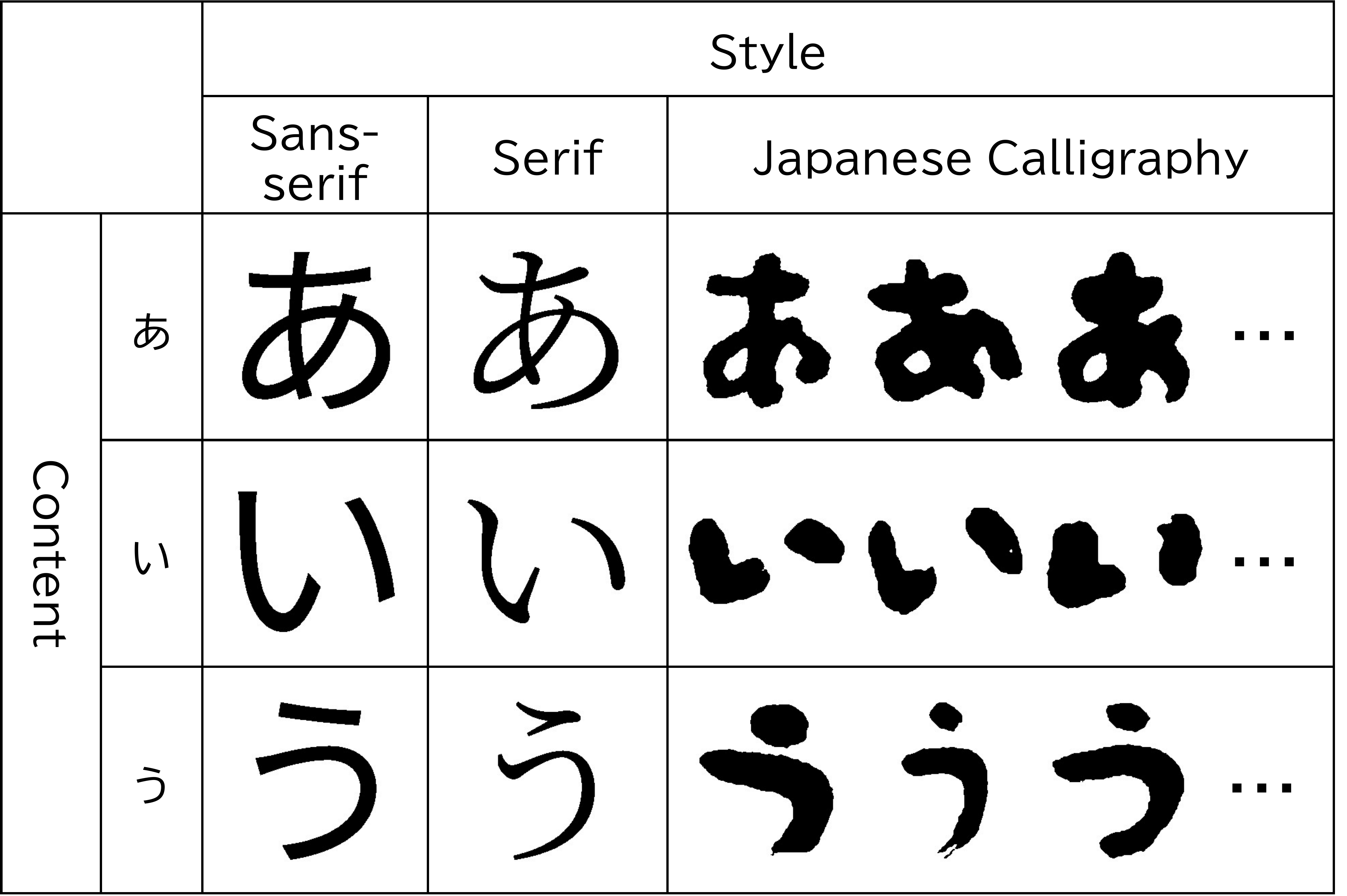}
\caption{Relationship among content, style, and variability in font and handwriting generation.}
\label{fig:variation}
\end{center}
\end{figure}

Font generation is a deterministic process; once the content and style are specified, the resulting glyph is uniquely determined. Conversely, handwriting generation is inherently non-deterministic. As illustrated in Fig.~\ref{fig:variation}, even when the same writer writes the identical content in a consistent style multiple times, the glyph shapes never match perfectly. In handwriting, glyph rendering involves an additional factor: \textbf{handwriting variability} (or simply \textbf{variability}), which reflects the natural variations in human writing behavior. This variability yields the natural, human-like appearance characteristic of hand-written text.

This handwriting variability presents two major challenges for conventional generative models. The first challenge is the instability of model training. When imitating a writer's style with limited samples, the model fails to capture the distribution of variability, leading to inconsistent glyph quality. The second challenge is the high cost of data collection. Synthesizing high-quality, realistic glyphs requires gathering handwritten samples that cover diverse variations.

We propose an approach that stabilizes generation quality while reducing dataset dependency. Rather than relying on pixel-level alignment with ground-truth images, our method incorporates human perceptual feedback into model training, bridging the gap between computational models and human cognitive judgment of handwriting naturalness. While writing large volumes of characters is labor-intensive, assessing pairwise preferences between generated images is straightforward for human evaluators. In this work, we apply Direct Preference Optimization (DPO), a variant of Reinforcement Learning from Human Feedback (RLHF), to the baseline model constructed by performing additional training on a small set of target handwriting glyphs.

Specifically, we employ Diffusion-DPO~\cite{2024:CVPR:Diffusion-DPO}, an extension of preference optimization to diffusion models. This framework directly integrates human preferences regarding subtle handwriting variability into the optimization process. Because the non-deterministic nature of handwriting naturally introduces variations in the outputs, humans can easily compare and evaluate their preferences among the generated samples.

Our key contributions are summarized as follows:
\begin{enumerate}
    \item We adapt Diffusion-DPO, originally designed for text-to-image models, into a training framework applicable to conditional glyph generation.
    \item We demonstrate that incorporating human preference feedback into handwriting generation models, starting from a small amount of real handwriting data, improves the readability of generated characters.
    \item We analyze how style complexity and handwriting variability influence DPO training, revealing the challenges of simultaneously optimizing both readability and style reproduction under a single preference criterion.
\end{enumerate}

\section{Related Work}

\subsection{Reinforcement Learning from Human Feedback}

Reinforcement Learning from Human Feedback (RLHF) optimizes an agent's policy using a reward model trained on human preference feedback. Historically, Christiano et al.~\cite{2017:NIPS:RLHF} introduced this approach to train deep reinforcement learning systems on complex goals using pairwise comparisons. Since then, RLHF has undergone extensive research. In particular, Proximal Policy Optimization (PPO)~\cite{2017:PPO}, which simplifies the implementation of Trust Region Policy Optimization (TRPO)~\cite{2015:ICML:TRPO}, has become a standard algorithm in LLM alignment pipelines.

Direct Preference Optimization (DPO)~\cite{2023:NeurIPS:DPO} addresses these limitations. Through mathematical reformulation, DPO bypasses explicit reward modeling and complex reinforcement learning loops, enabling direct optimization of the policy model via a single binary cross-entropy loss on preference data. This approach significantly simplifies and stabilizes the training process, accelerating the deployment of alignment techniques across various domains.

\subsection{Diffusion Models and Reinforcement Learning}

Fan and Lee~\cite{2023:ICML:Fan} first integrated policy gradient methods into Denoising Diffusion Probabilistic Models (DDPMs)~\cite{2020:NeurIPS:DDPM} to optimize the sampling process. Extending this approach to large-scale text-to-image models, frameworks like DPOK~\cite{2023:NeurIPS:DPOK} and DDPO~\cite{2024:ICLR:DDPO} treat the reverse diffusion process as a multi-step decision-making procedure. These methods successfully align models with human preferences, improving attributes such as aesthetic quality and prompt consistency.

Recently, Diffusion-DPO~\cite{2024:CVPR:Diffusion-DPO} adapted the DPO framework to diffusion models, optimizing parameters directly from pairwise preference data. This formulation eliminates the need for training a separate reward model, streamlining preference learning in generative image models.

\subsection{Glyph Generation}

Font generation, a subfield of glyph generation, aims to automate font design. Early efforts in this domain introduced GAN-based architectures such as zi2zi~\cite{2017:zi2zi}, which paved the way for subsequent GAN-based approaches~\cite{2025:GC-Font:BMVC,2021:ICMR:MLFont,2022:CVPR:FsFont}. Recent work increasingly leverages diffusion models. For instance, Diff-Font~\cite{2024:IJCV:Diff-Font} pioneered the application of diffusion models to font synthesis, while FontDiffuser~\cite{2024:AAAI:FontDiffuser} achieves effective style-content disentanglement and stroke preservation without requiring stroke annotations.

In contrast, handwriting generation research focuses on separating content, style, and intra-writer variation to model natural variability. Some approaches introduce noise to style features to ensure output diversity. For example, the Style-Disentangled Transformer (SDT)~\cite{2023:CVPR:Dai} separates style representation into a writer's overall style and content-dependent style, reporting that the former captures low-frequency image characteristics while the latter represents complex high-frequency style patterns. Building on this idea, One-DM~\cite{2024:ECCV:One-DM} implements the high-frequency variation hypothesis within a specific network architecture. Kuwata et al.~\cite{2024:BigComp:Kuwata} successfully captured and generated the pronounced variability inherent in Japanese calligraphy by separating style and content. GL-GAN~\cite{2025:COLING:GL-GAN} distinguishes between global style, representing overall traits such as stroke width and slant, and local style, focusing on glyph connections and individual character shapes.

Other methods, such as those proposed by Liu et al.~\cite{2023:ICSMD:Liu} and Kang et al.~\cite{2020:ECCV:GANwriting}, attempt to ensure diversity by injecting noise into style features or shuffling the permutation of reference image sequences. While these methods simulate handwriting variability by adding noise to the latent space, they rely on artificial perturbations. Rather than manually introducing synthetic noise to force diversity, our approach directly embeds human perceptual preference feedback into the optimization loop.

\section{Problem Formulation and Dataset}

\subsection{Problem Formulation}

This section defines the handwriting generation task and establishes the mathematical notation. As defined in Section~\ref{sec:intro}, characters are formulated using three primary components: \textbf{content}, \textbf{style}, and \textbf{glyph}. Our objective is to generate a new glyph image $y$ conditioned on a content reference image $x_c$ and a style reference image $x_s$.

Let $C = \{c_i\}_{i=1}^n$ be the set of character contents and $S = \{s_j\}_{j=1}^m$ be the set of styles. Defined over the image space $\mathcal{X}$, $R_c \subset \mathcal{X}$ denotes the set of images sharing a specific content $c \in C$ with arbitrary styles, and $R_s \subset \mathcal{X}$ denotes the set of images sharing a style $s \in S$ with arbitrary contents. Given a content reference image $x_c \in R_c$ and a handwritten style reference image $x_s \in R_s$, the model aims to synthesize a glyph $y$ that exhibits both the target content $c$ and the target style $s$.

\subsection{Dataset}

We collected handwriting data representing four styles and Japanese font data representing 14 styles. The font dataset serves to pre-train the model efficiently across diverse styles. For font data, we use glyphs covering all $n=6,524$ content classes in $C$ for each style. For handwriting data, we gather glyphs for a subset of $C$ that was practically obtainable.

\subsubsection{Japanese Handwriting Dataset}

The dataset contains four distinct styles: Etsuzan Tawara style, brush, marker, and pen. For the Etsuzan Tawara style, we extracted 1,479 glyphs (one image per content) from the public calligraphy works of Etsuzan Tawara~\footnote{\url{https://etsuzan.jp/}; permission was obtained from the calligrapher to use his works for this dataset.}, cropping each image with a tight bounding box. For the remaining three styles, we commissioned different writers to transcribe the identical set of 1,479 characters using a brush, marker, and pen, yielding a total of 5,916 handwritten glyph images.

\subsubsection{Japanese Font Dataset}

To cover visually distinct design features, we selected 14 fonts from Google Fonts~\footnote{\url{https://fonts.google.com/}} based on visual inspection. As a general selection rule, if multiple fonts shared the same typeface (e.g., Sans-serif or Serif), we selected only one representative font. We made exceptions for fonts with visual features that diverged significantly from typical designs. Table~\ref{table:use_font} lists the 14 selected fonts. Utilizing 6,524 glyphs per font (except for ReggaeOne-Regular, from which the 39 test content classes were excluded, yielding 6,485 glyphs) from the content set $C$, we obtained a total of 91,297 font images.

\begin{table}[tb]
  \centering
  \caption{List of the 14 Japanese fonts used in our dataset.}
  \label{table:use_font}
  \begin{tabular}{ll}
    \hline
    \hline
    DotGothic16-Regular &
    Stick-Regular \\
    KleeOne-Regular &
    ZenMaruGothic-Regular \\
    SourceHanSansJP-Regular &
    ZenKurenaido-Regular \\
    NewTegomin-Regular &
    HinaMincho-Regular \\
    YuseiMagic-Regular &
    SourceHanSerifJP-Regular \\
    ReggaeOne-Regular &
    KaiseiDecol-Regular \\
    ShipporiMincho-Regular &
    HachiMaruPop-Regular \\
    \hline
  \end{tabular}
\end{table}

\subsection{Image Preprocessing}

We resize each glyph image so that the longer side of its bounding box measures 236 pixels. We then add 10 pixels of padding to both ends of the longer side. Finally, we pad the shorter side to 256 pixels, standardizing all images to a square size of $256 \times 256$ pixels, which serves as the default resolution. Images are subsequently scaled to meet the input requirements of specific models. During preprocessing, all images are converted to grayscale, binarized via thresholding, and normalized to map pixel values to the range $[-1, 1]$.

\subsection{Test Data Selection}\label{sec:test_selection}

We selected four handwriting styles from our dataset as test targets: Etsuzan Tawara, brush, marker, and pen. As test content, we extracted 39 characters from the content set $C$ (6,524 characters), comprising 13 characters each from Hiragana, Katakana, and Kanji. These test content classes do not overlap with the 15 characters reserved for additional training (Section~\ref{sec:pretrain}). From the glyph set $R_s$ of each handwriting style, glyphs containing the selected test content classes served as candidate test samples. Among these candidates, we directly adopted unique glyphs, and randomly selected a single sample when multiple glyphs shared the identical content class.

We discarded any remaining duplicate candidate images to ensure they were excluded from both training and testing. All glyph images not included in the test candidates served as training data, where overlaps in content and style were permitted. Excluding glyphs containing the test content classes from the original 5,916 handwritten images (1,479 characters $\times$ 4 styles) resulted in 5,332 handwriting images (1,333 characters $\times$ 4 styles). Consequently, the final dataset consists of a training set containing 5,332 handwriting images and 91,297 font images, and a test set containing 156 handwriting images (39 characters $\times$ 4 styles).
\section{Training Glyph Generation Models Using DPO}

This section describes the training procedure for a single target handwriting style. In our experiments, we trained independent models for each of the four target handwriting styles using the identical pipeline.

While transcribing handwriting to compile a dataset is labor-intensive, providing preference feedback by comparing the relative quality of two generated glyphs is highly intuitive for human annotators. In this work, we first pre-train a generator model on Japanese font data and then perform additional training using a small handwriting dataset to construct the baseline model. We then collect pairwise preference annotations on the images synthesized by this baseline model. Using the compiled preference dataset, we apply Diffusion-DPO~\cite{2024:CVPR:Diffusion-DPO} to optimize the model to align with human perceptual judgments.

\begin{figure*}[tb]
    \centering
    \includegraphics[width=\linewidth]{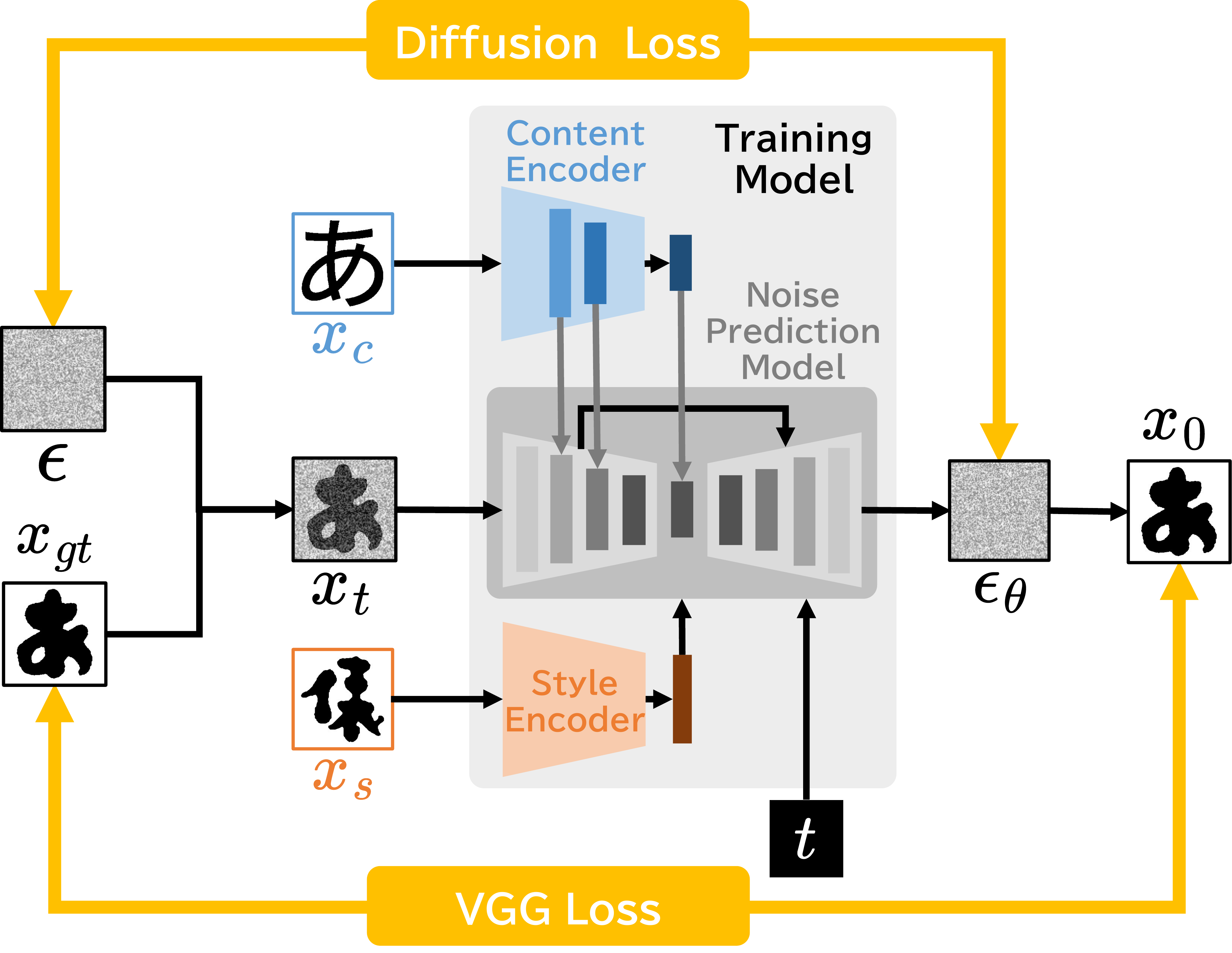}
    \caption{FontDiffuser-Lite}
    \label{fig:FontDiffuser_lite}
\end{figure*}

\subsection{Baseline Model Construction}\label{sec:pretrain}

We employ FontDiffuser-Lite, a lightweight variant of FontDiffuser~\cite{2024:AAAI:FontDiffuser}, as our generator model. Fig.~\ref{fig:FontDiffuser_lite} illustrates the architecture of FontDiffuser-Lite, which consists of a content encoder, a style encoder, and a noise prediction model. Both pre-training and additional training utilize triplet data consisting of a content reference image $x_c$, a style reference image $x_s$, and a ground-truth image $x_{gt}$. Among the 14 font styles in our Japanese font dataset, SourceHanSansJP-Regular (a Gothic font) is selected as the source style for the content reference image $x_c$.

In pre-training, glyphs from the remaining 13 font styles serve as style reference images $x_s$ and ground-truth images $x_{gt}$. In additional training, we select one specific handwriting style as the target. We extract two glyph samples per content class to use as style reference images $x_s$ and ground-truth images $x_{gt}$. For this purpose, we selected 15 target content classes (five each from Hiragana, Katakana, and Kanji): ``\Ja{あ, か, な, わ, を, ウ, キ, タ, ツ, ル, 敗, 意, 起, 談, 年}.'' Consequently, additional training utilizes a total of 30 glyphs (two samples per content class).

The content reference image $x_c$ and style reference image $x_s$ are fed into the content encoder and style encoder, respectively. The extracted content features are incorporated into the encoder and bridge layers of the noise prediction model, while the style features are projected into the noise prediction model's layers via cross-attention. Our generation model builds on Denoising Diffusion Probabilistic Models (DDPMs)~\cite{2020:NeurIPS:DDPM}. The forward diffusion process adds Gaussian noise to the target distribution $x_0 \sim q(x_0)$ step-by-step. Let $t = 1, \dots, T$ be the diffusion steps, and $\beta_1 < \dots < \beta_T$ be variance schedules. Defining $\alpha_t = 1 - \beta_t$ and $\bar{\alpha}_t = \prod^t_{i=1} \alpha_i$, the noisy image $x_t$ at step $t$ is formulated as follows:
\begin{align}
x_t
&=\sqrt{1-\beta_t}x_{t-1}+\sqrt{\beta_t}\epsilon_{t-1},~\epsilon_{t-1}\sim\mathcal{N}(0,I)\\
&=\sqrt{\bar{\alpha}_t}x_0+\sqrt{1-\bar{\alpha}_t}\epsilon,~\epsilon\sim\mathcal{N}(0,I)
\label{eq:diffusion_process}
\end{align}
Here, $\epsilon$ represents the added noise. We feed $x_t$, generated by adding noise to $x_{gt}$ at step $t$, into the noise prediction model to estimate the added perturbation. Reversing this diffusion process based on the predicted noise reconstructs $\hat{x}_0$, which estimates the original ground-truth image $x_{gt}$ (i.e., $x_0$).

Training minimizes a joint loss composed of two terms. The first term is the diffusion loss $\mathcal{L}_{diff}$, which evaluates the mean squared error (MSE) between the added noise $\epsilon$ and the output of the noise prediction model $\epsilon_\theta$:
\begin{equation}
\mathcal{L}_{diff}=\lVert\epsilon-\epsilon_\theta(x_t,t,x_c, x_s)\rVert_2^2
\end{equation}
The second term is the perceptual consistency loss (VGG loss) $\mathcal{L}_{cp}$, which measures the similarity between the reconstructed image $x_0$ and the ground-truth image $x_{gt}$. It computes the average difference in feature maps across layers $l = 1 \dots L$ of a pre-trained VGG network, where $L$ denotes the total number of layers:
\begin{equation}
\mathcal{L}_{cp} = \sum_{l=1}^{L} \|\mathcal{VGG}_l(x_0) - \mathcal{VGG}_l(x_{gt})\|
\end{equation}

\subsection{Preference Feedback Collection and Dataset Construction}\label{sec:dpo_anotation}

We construct the preference dataset using the baseline model. For a given pair of content reference image $x_c$ and style reference image $x_s$, we independently generate two glyph images. A human annotator compares the outputs, labeling the superior image as the winner $x_w$ and the inferior one as the loser $x_l$, to assemble quadruplet samples of $(x_c, x_s, x_w, x_l)$.

The content classes generated for the preference dataset align with the character vocabulary present in the training set of the target style (1,333 real images, detailed in Section~\ref{sec:test_selection}). We use the SourceHanSansJP-Regular glyph of the matching character class as the content reference image $x_c$. For the style reference image, we randomly sample from the 30 glyphs allocated for additional training in Section~\ref{sec:pretrain}. Based on these settings, we generated two images for each of the 1,333 input conditions (totaling 2,666 generated images) and subjected them to preference annotation.

Annotators evaluated the outputs using the following prioritized criteria:
\begin{enumerate}
    \item The output must be free of peripheral artifacts (e.g., unintended dots or strokes), and the background must remain clean.
    \item The length and placement of strokes must yield a legible, structurally correct character.
    \item The generated glyph must capture target style features such as line width variation, rounded corners, and ink characteristics (e.g., bleeding or fading).
\end{enumerate}
If the two generated glyphs were of equal quality, making preference judgment impossible, they were treated as excluded pairs and discarded from the dataset. To maintain consistency and eliminate subjective bias, the first author conducted all annotations and exclusions alone.

Table~\ref{table:anotation} shows examples of preference annotations. Based on the evaluation, we labeled excluded pairs as 0, pairs where image 1 was preferred (where image 1 becomes $x_w$, and image 2 becomes $x_l$) as 1, and pairs where image 2 was preferred (where image 2 becomes $x_w$, and image 1 becomes $x_l$) as 2, forming the final preference dataset.

\begin{table}[tb]
  \caption{Examples of preference annotation labelling (0, 1, and 2).}
  \label{table:anotation}
  \centering
  \begin{tabular}{ccc|cc|c}
    \hline \hline
    Content & Style & Real & Generation1 & Generation2 & Label \\
    \hline
    \begin{minipage}{0.14\columnwidth}
      \centering
      \scalebox{0.45}{\includegraphics{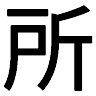}}
    \end{minipage} &
    \begin{minipage}{0.14\columnwidth}
      \centering
      \scalebox{0.45}{\includegraphics{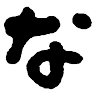}}
    \end{minipage} &
    \begin{minipage}{0.14\columnwidth}
      \centering
      \scalebox{0.45}{\includegraphics{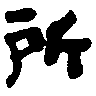}}
    \end{minipage} &
    \begin{minipage}{0.14\columnwidth}
      \centering
      \scalebox{0.45}{\includegraphics{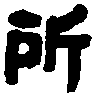}}
    \end{minipage} &
    \begin{minipage}{0.14\columnwidth}
      \centering
      \scalebox{0.45}{\includegraphics{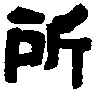}}
    \end{minipage} &
    0 \\
    \hline
    \begin{minipage}{0.14\columnwidth}
      \centering
      \scalebox{0.45}{\includegraphics{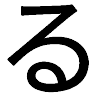}}
    \end{minipage} &
    \begin{minipage}{0.14\columnwidth}
      \centering
      \scalebox{0.45}{\includegraphics{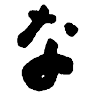}}
    \end{minipage} &
    \begin{minipage}{0.14\columnwidth}
      \centering
      \scalebox{0.45}{\includegraphics{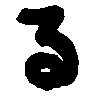}}
    \end{minipage} &
    \begin{minipage}{0.14\columnwidth}
      \centering
      \scalebox{0.45}{\includegraphics{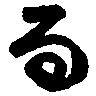}}
    \end{minipage} &
    \begin{minipage}{0.14\columnwidth}
      \centering
      \scalebox{0.45}{\includegraphics{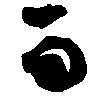}}
    \end{minipage} &
    1 \\
    \hline
    \begin{minipage}{0.14\columnwidth}
      \centering
      \scalebox{0.45}{\includegraphics{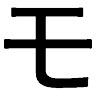}}
    \end{minipage} &
    \begin{minipage}{0.14\columnwidth}
      \centering
      \scalebox{0.45}{\includegraphics{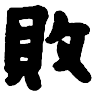}}
    \end{minipage} &
    \begin{minipage}{0.14\columnwidth}
      \centering
      \scalebox{0.45}{\includegraphics{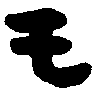}}
    \end{minipage} &
    \begin{minipage}{0.14\columnwidth}
      \centering
      \scalebox{0.45}{\includegraphics{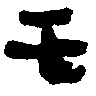}}
    \end{minipage} &
    \begin{minipage}{0.14\columnwidth}
      \centering
      \scalebox{0.45}{\includegraphics{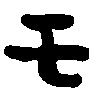}}
    \end{minipage} &
    2 \\
    \hline
  \end{tabular}
\end{table}

\begin{figure*}[tb]
    \centering
    \includegraphics[width=\linewidth]{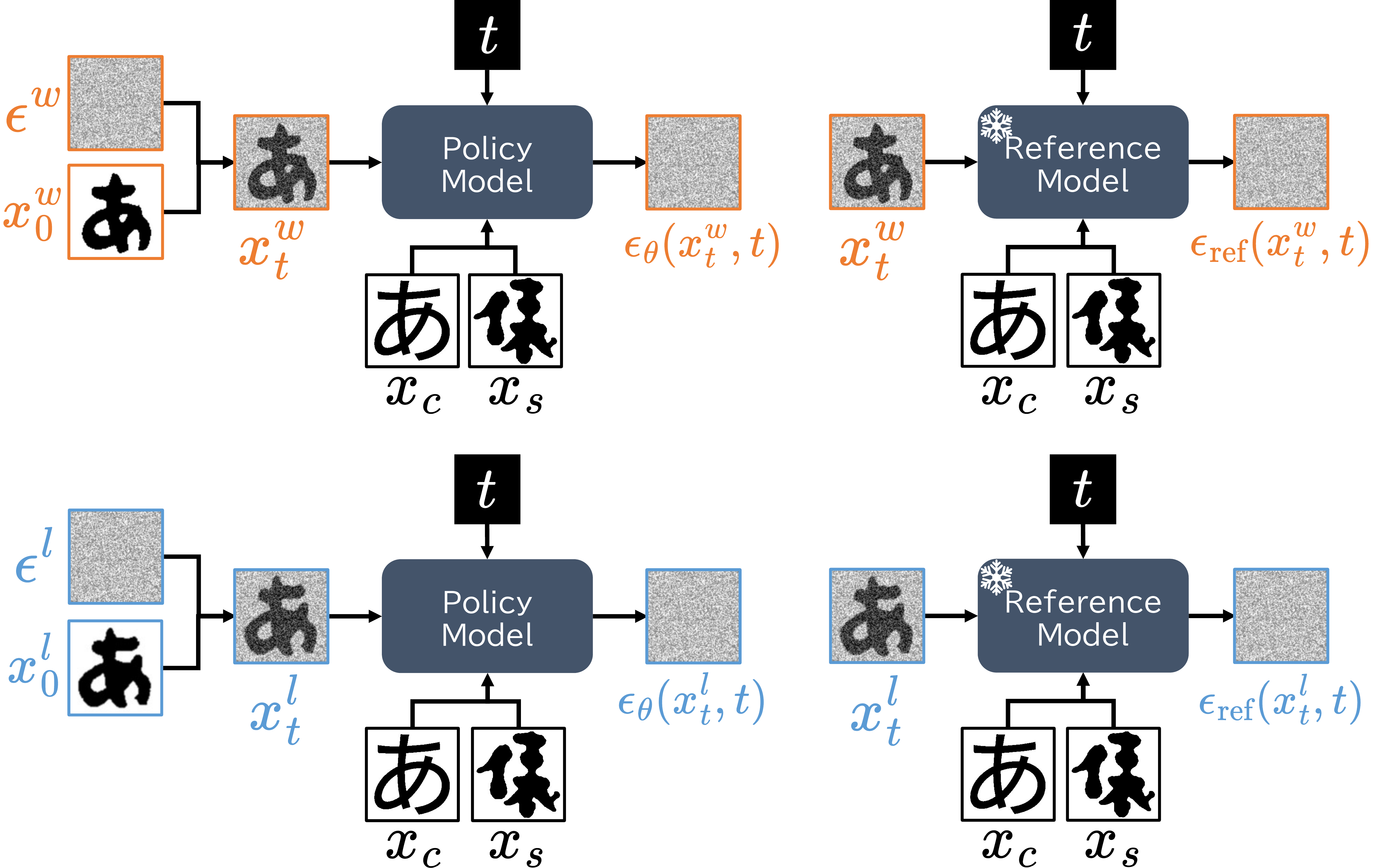}
    \caption{Overview of inputs and outputs for the policy and reference models during the DPO training process.}
    \label{fig:DPO}
\end{figure*}

\subsection{Optimization via DPO}

Diffusion-DPO~\cite{2024:CVPR:Diffusion-DPO} training employs two neural networks: the policy model, which undergoes parameter updates, and the reference model, which remains frozen. Fig.~\ref{fig:DPO} illustrates the inputs and outputs of these models during training. At the start of the optimization, both the policy model and the reference model are initialized with the identical weights of the baseline model. During optimization, the reference model's parameters remain frozen.

Let $x_0^w$ and $x_0^l$ denote the winner and loser images from the preference dataset, respectively. We denote the random Gaussian noise applied to these images as $\epsilon^w$ and $\epsilon^l$. The resulting noisy images at diffusion step $t$ are represented as $x_t^w$ and $x_t^l$.

Using the policy model and the reference model, we estimate the added noise for both the noisy winner image $x_t^w$ and the noisy loser image $x_t^l$. Specifically, conditioned on the content reference image $x_c$ and style reference image $x_s$, we predict four distinct noise representations: the policy model's estimates $\epsilon_\theta(x_t^w, t)$ and $\epsilon_\theta(x_t^l, t)$, and the reference model's estimates $\epsilon_\text{ref}(x_t^w, t)$ and $\epsilon_\text{ref}(x_t^l, t)$.

We optimize the policy model using a weighted sum of three loss functions. The primary objective is the Diffusion-DPO loss $\mathcal{L}_{dpo}(\theta)$, formulated as follows:
\begin{equation}
\begin{aligned}
    \mathcal{L}_{dpo}(\theta) = & -\mathbf{E}_{t, \epsilon^w, \epsilon^l} \log \sigma ( -\beta T \omega(\lambda_t) ( \\
    & \|\epsilon^w - \epsilon_\theta(x_t^w, t)\|^2 - \|\epsilon^w - \epsilon_{\text{ref}}(x_t^w, t)\|^2 \\ 
    & - (\|\epsilon^l - \epsilon_\theta(x_t^l, t)\|^2 - \|\epsilon^l - \epsilon_{\text{ref}}(x_t^l, t)\|^2) ) )
\end{aligned}
\end{equation}
Here, $\beta$ is a scaling regularization parameter, $T$ denotes the total number of diffusion steps, $\lambda_t$ represents the signal-to-noise ratio (SNR) at step $t$, $\omega$ is a weighting function, and $\sigma$ is the sigmoid function. This loss function computes the squared error between the estimated noise and the target noise (noise prediction error) for the four predictions. For both the winner and loser images, the relative improvement of the policy model is calculated by subtracting the prediction error of the reference model from that of the policy model. Optimization updates the policy model weights to maximize the relative improvement on the winner images while minimizing it on the loser images. The secondary and tertiary terms are the diffusion loss and the perceptual consistency loss (VGG loss), defined identically to those in Section~\ref{sec:pretrain}, but computed exclusively using the winner images from the preference dataset.

\section{Experiments and Evaluation}

To verify the effectiveness of the proposed DPO-based training, we compared the outputs of our model with those of the baseline model. The objective of these experiments is to evaluate whether preference-based optimization can compensate for data scarcity and improve synthesis quality when the number of target glyphs for additional training is extremely limited.

\subsection{Experimental Settings}

We first pre-trained FontDiffuser-Lite on the Japanese font dataset. Optimization utilized AdamW with hyperparameters set to $\beta_1 = 0.9$, $\beta_2 = 0.999$, $\epsilon = 10^{-8}$, and $\text{weight decay} = 10^{-2}$. We set the learning rate to $0.0001$, employing a linear decay schedule with $10,000$ warm-up steps. The probability of unconditional conditioning ($\emptyset$) in the classifier-free guidance strategy was set to $0.1$. Pre-training ran for 30 epochs. For sampling, we set the guidance scale to $s=7.5$ and performed 20 denoising steps using the DPM-Solver++ sampler. The sampled images were binarized via thresholding to yield the final outputs.

Using the pre-trained weights, we performed additional training for each target handwriting style. The optimization settings differed from pre-training in that we fixed the learning rate to $0.00001$ using a constant schedule. Additional training ran for 300 epochs. The resulting models are defined as the baseline model, which we used to generate samples for constructing the preference dataset.

To collect preference data, we generated images focusing on the target styles. As described in Section~\ref{sec:dpo_anotation}, we synthesized samples under 1,333 conditions and conducted preference annotation. The process yielded 30 valid preference pairs for Etsuzan Tawara, 134 for brush, 122 for marker, and 177 for pen. Using these datasets, we fine-tuned the baseline models via Diffusion-DPO to obtain the final target models. For DPO training, the learning rate was set to $0.00001$ with a constant schedule. We did not apply the classifier-free guidance strategy during DPO training. Because the size of the preference dataset varied across styles, we standardized the training duration by step counts rather than epochs, conducting exactly 200 optimization steps for all styles. We executed this pipeline independently for each of the four handwriting styles.

\begin{table*}[tb]
  \caption{Visual comparison of generated glyphs from each model against the ground-truth (Real) images.}
  \label{table:glyph_generation_result}
  \centering
  \begin{tabular}{ccc|ccc|ccc|ccc}
    \hline \hline
    \multicolumn{3}{c|}{Etsuzan Tawara} & \multicolumn{3}{c|}{Brush} & \multicolumn{3}{c|}{Marker} & \multicolumn{3}{c}{Pen} \\
    \hline
    Real & Baseline & DPO & Real & Baseline & DPO & Real & Baseline & DPO & Real & Baseline & DPO\\
    \hline
    \begin{minipage}{0.06\columnwidth}
      \centering
      \scalebox{0.22}{\includegraphics{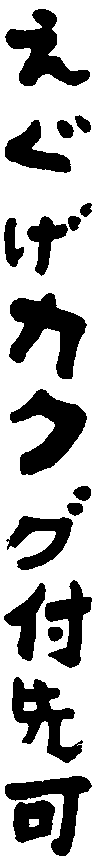}}
    \end{minipage} &
    \begin{minipage}{0.06\columnwidth}
      \centering
      \scalebox{0.22}{\includegraphics{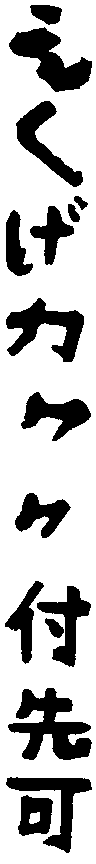}}
    \end{minipage} &
    \begin{minipage}{0.06\columnwidth}
      \centering
      \scalebox{0.22}{\includegraphics{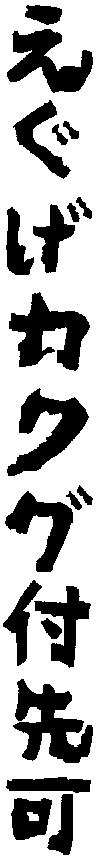}}
    \end{minipage} &
    \begin{minipage}{0.06\columnwidth}
      \centering
      \scalebox{0.22}{\includegraphics{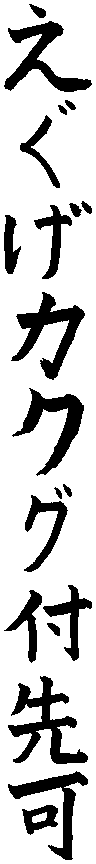}}
    \end{minipage} &
    \begin{minipage}{0.06\columnwidth}
      \centering
      \scalebox{0.22}{\includegraphics{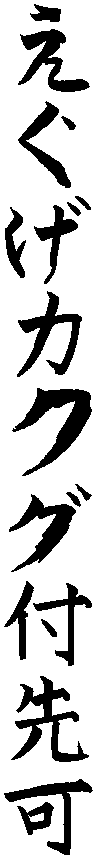}}
    \end{minipage} &
    \begin{minipage}{0.06\columnwidth}
      \centering
      \scalebox{0.22}{\includegraphics{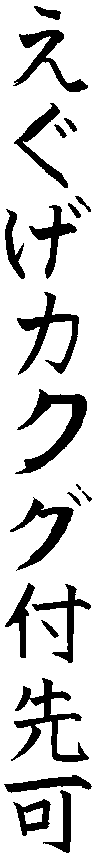}}
    \end{minipage} &
    \begin{minipage}{0.06\columnwidth}
      \centering
      \scalebox{0.22}{\includegraphics{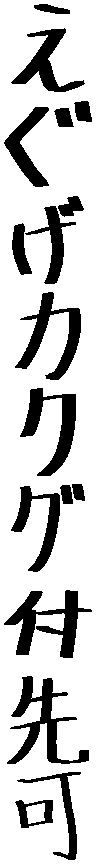}}
    \end{minipage} &
    \begin{minipage}{0.06\columnwidth}
      \centering
      \scalebox{0.22}{\includegraphics{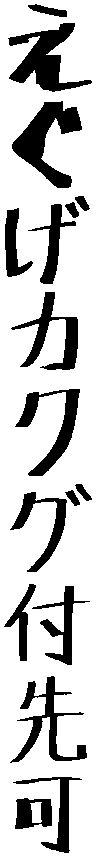}}
    \end{minipage} &
    \begin{minipage}{0.06\columnwidth}
      \centering
      \scalebox{0.22}{\includegraphics{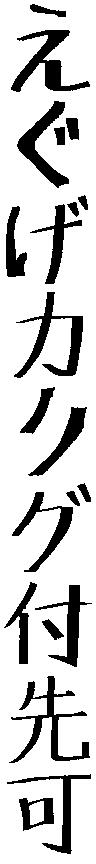}}
    \end{minipage} &
    \begin{minipage}{0.06\columnwidth}
      \centering
      \scalebox{0.22}{\includegraphics{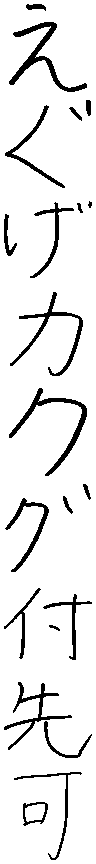}}
    \end{minipage} &
    \begin{minipage}{0.06\columnwidth}
      \centering
      \scalebox{0.22}{\includegraphics{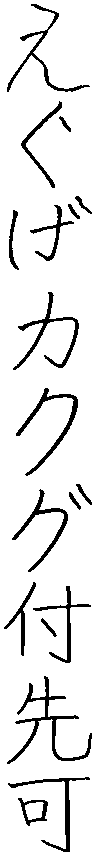}}
    \end{minipage} &
    \begin{minipage}{0.06\columnwidth}
      \centering
      \scalebox{0.22}{\includegraphics{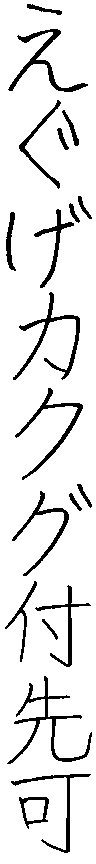}}
    \end{minipage} \\
    \hline
  \end{tabular}
\end{table*}

\subsection{Evaluation Metrics}

We evaluate the quality of the generated images using both automatic evaluation metrics and human evaluation. For automatic evaluation, we adopt Learned Perceptual Image Patch Similarity (LPIPS)~\cite{2018:CVPR:LPIPS} and Fr\'echet Inception Distance (FID)~\cite{2017:NIPS:FID}. LPIPS measures the perceptual similarity between images based on feature representations extracted from deep learning networks. FID quantifies the similarity between the distribution of real images and generated images.

For human evaluation, we conducted a subjective user study with two native Japanese-speaking university students to evaluate style reproduction and content readability. The evaluators did not possess professional expertise in calligraphy or design, providing a general public perspective. The study followed a double-blind protocol in which the generation method for each image was hidden, and images were presented in a randomized order. Evaluators rated each sample on a 5-point Likert scale (from 1: extremely poor to 5: extremely good) across the two evaluation items.

Style reproduction assessed whether the visual characteristics of the target style were successfully imitated. Specifically, we provided style exemplars (real images) and asked evaluators to focus on style-specific features, such as brush strokes, pen traits, and fading, even if the underlying character skeleton was deformed. Content readability evaluated whether the synthesized glyph was legible and recognizable as a valid character. Evaluators rated readability focusing solely on whether the character's content was clearly identifiable, regardless of style consistency. To ensure objectivity, the two evaluators worked independently without consultation. While the ground-truth images were displayed as references, ratings were confined strictly to the generated samples.

\begin{table*}[tb]
  \caption{Quantitative results under automatic evaluation metrics and human evaluation.}
  \label{table:glyph_generation_quantitative}
  \centering
  \begin{tabular}{ll|cc|cc}
    \hline
    \hline
     & & \multicolumn{2}{c|}{Automatic evaluation} & \multicolumn{2}{c}{Human evaluation} \\
    \hline
    Style & Model  & LPIPS$\downarrow$ & FID$\downarrow$ & Style$\uparrow$ & Content$\uparrow$ \\
    \hline
    \multirow{2}{*}{Etsuzan Tawara}
        & Baseline & \textbf{0.2622} & \textbf{125.5543} & \textbf{4.31} & 3.96 \\
        & DPO & 0.2649 & 129.5528 & 3.87 & \textbf{4.08} \\
    \hline
    \multirow{2}{*}{Brush}
        & Baseline & \textbf{0.2213} & \textbf{97.1554} & \textbf{4.12} & 4.44 \\
        & DPO & 0.2275 & 99.4174 & 3.83 & \textbf{4.55} \\
    \hline
    \multirow{2}{*}{Marker}
        & Baseline & 0.2200 & 93.1845 & 4.03 & 4.49 \\
        & DPO  & \textbf{0.2160} & \textbf{91.3261} & \textbf{4.12} & \textbf{4.73} \\
    \hline
    \multirow{2}{*}{Pen}
        & Baseline & \textbf{0.1977} & 74.3895 & 4.65 & 4.65 \\
        & DPO & 0.1999 & \textbf{71.3685} & \textbf{4.68} & \textbf{4.77} \\
    \hline
  \end{tabular}
\end{table*}

\subsection{Experimental Results}

Table~\ref{table:glyph_generation_result} presents the qualitative synthesis results of the baseline and DPO-tuned models, with real images shown for reference. Across all styles, DPO-based optimization generated glyphs with stable structural skeletons and fewer structural collapses compared to the baseline. This observation indicates that training with human preference feedback successfully encourages the model to prioritize shapes that are legible to humans. Style preservation, however, exhibited style-dependent tendencies. For styles with relatively low variability, such as marker and pen, stylistic features were successfully preserved without significant distortion. Conversely, styles with complex textures and highly diverse shapes, such as Etsuzan Tawara and brush, exhibited noticeable stylistic degradation or collapses.

Table~\ref{table:glyph_generation_quantitative} summarizes the quantitative evaluation results. Regarding automatic metrics, DPO tuning yielded no numeric improvement in either LPIPS or FID for the Etsuzan Tawara and brush styles. In contrast, the marker style showed improvements in both metrics after DPO tuning. For the pen style, the results were mixed: the baseline model performed better in LPIPS, whereas the DPO-tuned model excelled in FID.

Regarding human evaluation, we measured the inter-rater reliability between the two evaluators using Cohen's weighted kappa coefficient, yielding $\kappa = 0.530$, which indicates moderate agreement. Comparing average scores, the DPO-tuned model outperformed the baseline in content readability across all styles. This improvement suggests that DPO training successfully stabilizes character structures and enhances readability. The average scores for style reproduction differed by style. For the marker and pen styles, style reproduction scores improved, indicating that DPO contributes to both readability enhancement and style preservation. For the Etsuzan Tawara and brush styles, style reproduction scores decreased.

\subsection{Discussion}

Our experiments confirm that DPO training generally improves the structural accuracy of generated glyphs. However, in complex styles with significant handwriting variability (e.g., the brush style), the model tends to trade off fine style-specific features to prioritize structural correctness.

Two factors likely explain this phenomenon. The first factor is style complexity. Dynamic style features such as ink fading, stroke-width variation, and complex contours (prominent in brush and Etsuzan Tawara styles) tend to be processed as noise during optimization. Consequently, the model converges toward generating simplified glyph representations that compromise stylistic fidelity. The brush style suffered a decline in style reproduction despite having a relatively large number of preference pairs (134 pairs), suggesting that style complexity itself, rather than data volume, exerts a dominant influence.

The second factor is the scarcity of preference annotations. For the Etsuzan Tawara style, the output quality of the baseline model was highly unstable. Many generated pairs failed to meet the evaluation baseline, resulting in their exclusion during annotation and reducing the number of valid preference samples to only 30. This limited data volume was likely insufficient to train the model to capture complex style characteristics while maintaining structural accuracy.

\section{Conclusion and Future Work}

This paper proposed a handwriting generation framework leveraging Diffusion-DPO. Unlike font generation, which is a deterministic process, handwriting generation is characterized by the non-deterministic nature where glyph shapes vary even under identical style and content conditions. While this handwriting variability is essential for natural human expression, it causes training instability and increases dataset acquisition costs for conventional generative models.

To address these challenges, we integrated human preference feedback directly into the optimization process by evaluating pairs of generated glyphs. This approach aims to improve generation quality while keeping dataset collection costs low. Experimental evaluations demonstrated that incorporating preference learning is highly effective for improving character readability. However, the current settings proved insufficient to fully preserve the details of complex style-specific features.

We identify two key directions for future work.

The first direction involves improving the efficiency, quality, and volume of preference annotations. In our experiments, many generated pairs exhibited minor visual differences, forcing us to discard them from the dataset as excluded pairs. Future work will explore sampler modifications and data augmentation techniques to generate diverse pairs containing clear distinctions, such as high-quality failure cases contrasted with success cases. Building systems that efficiently accumulate high-quality preference data remains essential.

The second direction is developing optimization methods for multiple preference criteria. In the current pipeline, human annotators subjectively prioritize multiple criteria, such as structural correctness and style reproduction, to assign a single preference label. Because these evaluation criteria can conflict, future work should define each metric as a distinct preference dimension and incorporate multi-objective optimization methods to train the model simultaneously across these criteria.

\begin{credits}
\subsubsection{\ackname} This work was supported by JSPS KAKENHI Grant Numbers JP25K03229, JP25K03228, and JP24K03228.

\subsubsection{\discintname}
The authors have no competing interests to declare that are relevant to the content of this article.

\subsubsection{Generative AI Usage Disclosure.}
During the preparation of this work, the authors used Gemini 3.5 Flash in order to improve the English phrasing and readability of the manuscript. After using this tool, the authors reviewed and edited the content as needed and take full responsibility for the content of the publication.
\end{credits}

\bibliographystyle{splncs04}
\bibliography{reference}

\end{document}